%% file: main.tex
\documentclass[10pt,twocolumn,letterpaper]{article}

\usepackage[pagenumbers]{iccv}

\input{preamble}

\definecolor{iccvblue}{rgb}{0.21,0.49,0.74}
\usepackage[pagebackref,breaklinks,colorlinks,allcolors=iccvblue]{hyperref}

\title{\textit{Bring Your Rear Cameras} \\ for Egocentric 3D Human Pose Estimation} 

\author{
Hiroyasu Akada\and
Jian Wang\and
Vladislav Golyanik\and
Christian Theobalt \and
Max Planck Institute for Informatics, SIC 
}

\begin{document}

\twocolumn[{
\maketitle

\renewcommand\twocolumn[1][]{#1}%
\begin{center}
    \captionsetup{type=figure}
    \includegraphics[width=\linewidth]{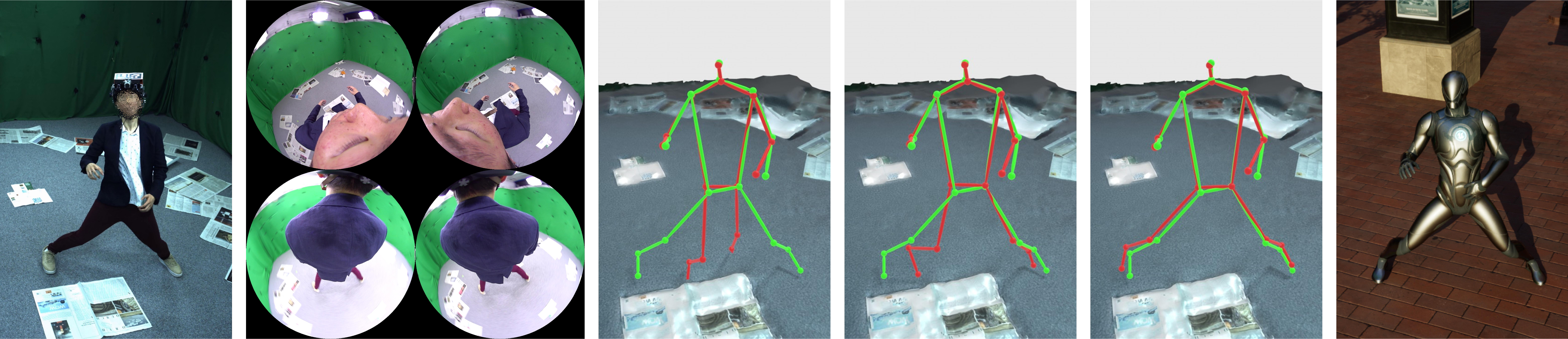}
    \vspace{-7mm}
    \caption*{\footnotesize{
        \,\,
        (a) Reference view
        \,\,\,\,\,\,\,\,\,
        (b) \begin{tabular}[l]{@{}c@{}}Egocentric fisheye images \\ front and rear views\end{tabular}
        \,\,\,\,\,\,\,\,
        (c) \begin{tabular}[l]{@{}c@{}} EgoPoseFormer \\ 2 front views \end{tabular}
        \,
        (d) \begin{tabular}[l]{@{}c@{}} EgoPoseFormer \\ 2 front / 2 rear views \end{tabular}
        (e) \begin{tabular}[l]{@{}c@{}} Our method \\ 2 front / 2 rear views\end{tabular}
        (f) \begin{tabular}[l]{@{}c@{}}Character animation \\ (future application) \end{tabular}
    }}
    \vspace{-3mm}
    \captionof{figure}{\textbf{3D human pose estimation of EgoPoseFormer~\cite{yang2024egoposeformer} and our method with a possible future application of 3D virtual character animation.} Predictions and ground truths are colored in red and green, respectively. See our video for dynamic visualizations.
    } 
    \label{fig:teaser}
\end{center}
}]

\input{sec/0_abstract}

\input{sec/1_intro}
\input{sec/2_related_work}

\input{sec/3_dataset}
\input{sec/4_method}

\input{sec/5_experiment}

\input{sec/6_conclusion}

{
    \small
    \bibliographystyle{ieeenat_fullname}
    \bibliography{main}
}

\maketitlesupplementary

\def\thesection{\Alph{section}}
\setcounter{section}{0}
\setcounter{figure}{6}
\setcounter{table}{5}
\setcounter{equation}{12}

\input{sec/X_suppl}

\end{document}

%% file: preamble.tex
\usepackage{times}
\usepackage{epsfig}
\usepackage{graphicx}
\usepackage{amsmath}
\usepackage{amssymb}

\usepackage{here}
\usepackage{caption}
\usepackage{multirow}
\usepackage{chngpage}
\usepackage{arydshln}
\usepackage{gensymb}
\usepackage{url}
\usepackage{color}
\usepackage{comment}
\usepackage{enumitem}
\usepackage{xspace}
\usepackage[normalem]{ulem}
\usepackage{soul}
\usepackage{hhline}

\usepackage[dvipsnames]{xcolor}

\usepackage[accsupp]{axessibility}  % Improves PDF readability for those with disabilities.

\usepackage{pifont}
\newcommand{\cmark}{\textcolor{teal}{\ding{51}}}%
\newcommand{\xmark}{\textcolor{red}{\ding{55}}}%
\def\eg{\emph{e.g.,}\xspace}
\def\ie{\emph{i.e.,}\xspace}

\def\etal{\emph{et al.}\xspace}

\definecolor{darkorange}{rgb}{1.0, 0.45, 0.0}

%% file: sec/0_abstract.tex
\begin{abstract} 
Egocentric 3D human pose estimation has been actively studied using cameras installed in front of a head-mounted device (HMD). 
While frontal placement is the optimal and the only option for some tasks, such as hand tracking, it remains unclear if the same holds for full-body tracking due to self-occlusion and limited field-of-view coverage. 
Notably, even the state-of-the-art methods often fail to estimate accurate 3D poses in many scenarios, such as when HMD users tilt their heads upward---a common motion in human activities. 
A key limitation of existing HMD designs is their neglect of the back of the body, despite its potential to provide crucial 3D reconstruction cues.
Hence, this paper investigates the usefulness of rear cameras for full-body tracking. 
We also show that simply adding rear views to the frontal inputs is not optimal for existing methods due to their dependence on individual 2D joint detectors without effective multi-view integration.
To address this issue, we propose a new transformer-based method that refines 2D joint heatmap estimation with multi-view information and heatmap uncertainty, thereby improving 3D pose tracking.
Also, we introduce two new large-scale datasets, Ego4View-Syn and Ego4View-RW, for a rear-view evaluation.
Our experiments show that the new camera configurations with back views provide superior support for 3D pose tracking compared to only frontal placements.  
The proposed method achieves significant improvement over the current state of the art (${>}10\%$ on MPJPE). 
\textbf{The source code, trained models, and datasets are available on our project page\footnote{\url{https://4dqv.mpi-inf.mpg.de/EgoRear/}}.} 

\end{abstract}

%% file: sec/1_intro.tex
\section{Introduction} 
\label{sec:intro}

Egocentric 3D human pose estimation has been explored with cameras installed in front of head-mounted devices (HMDs)~\cite{xu2019mo2cap2, tome2019xr, Tom2020SelfPose3E, zhang2021automatic, wang2021estimating, park2023domain, wang2022estimating, jiang2021egocentric, yuan2019ego, luo2021dynamics, Liu2023, Millerdurai_EventEgo3D_2024, wang2024egocentric, rhodin2016egocap, zhao2021egoglass, hakada2022unrealego, kang2023ego3dpose, kang2024egotap, hakada2024unrealego2, yang2024egoposeformer, eventegoplusplus}.
This design choice is driven by several factors, particularly the specific tasks for which HMDs are primarily optimized.
For instance, front cameras are the most practical (and only) option for hand tracking~\cite{li2013model, bambach2015lending, liu2024single, EgoPressure, Grauman_etal_2022_CVPR, Grauman_2024_CVPR}. 
However, it remains unclear if this camera placement is the best choice for all tasks, such as 3D full-body tracking.
Notably, Apple Vision Pro~\cite{applevisionpro}, a high-end HMD, has many sensors (eight cameras and more) placed only in the front for various tasks, yet does not officially offer full-body tracking (potentially due to low accuracy with only frontal inputs).
We also observe that due to self-occlusion and limited field-of-view coverage with the existing HMD designs, even the state-of-the-art method~\cite{yang2024egoposeformer} fails to estimate accurate 3D poses in many scenarios, such as when HMD users tilt their heads upward (Fig~\ref{fig:teaser}-(c))---a common human motion when exercising.

One solution to this challenge is to utilize body-mounted sensors (\eg~IMUs~\cite{jiang2022avatarposer, EgoLocate2023, armani2024ultrainertialposer, jiang2024egoposer, Rozumnyi_2025_WACV}) attached to the user’s limbs independently of HMDs (\eg~Meta Quest~\cite{metaquest}).
This necessitates additional sensor fusion and communication between the HMD and external devices.
Another solution involves integrating cameras in non-traditional locations on the HMD. 
In particular, rear cameras, an under-explored element in existing HMD designs, could offer valuable cues for 3D body tracking. 
(Beyond that, such cameras could also enhance avatar reconstruction, balance HMD weight distribution, and improve environmental sensing to prevent collisions with individuals approaching from behind.)

This paper explores the usefulness of rear cameras along with frontal ones for egocentric 3D body tracking.
We observe that naive fusion of rear views with frontal inputs does not always achieve optimal estimation for existing methods (Fig.~\ref{fig:teaser}-(d)) or can degrade accuracy (Fig.~\ref{fig:qualitaitve_result_pose3d}-(bottom right), 5:58$\sim$ in our supplementary video). 
This is due to their dependence on 2D joint estimators without effective multi-view integration.
As a result, self-occlusion and missing body parts---particularly in rear views---lead to inaccurate 2D joint detection and accuracy drops in 3D body tracking.

To this end, we propose a new transformer-based method that refines 2D joint heatmaps with multi-view context, improving 3D human pose estimation (Fig.~\ref{fig:teaser}-(e)). 
Our method is tailored for front and rear view data, based on the implicit assumption that front and rear views could complement each other due to the symmetrical nature of the human body.
That is, unreliable rear-view heatmaps could be improved by reliable front-view heatmaps and vice versa.
We first estimate 2D joint heatmaps from each view via a 2D joint detector (Sec.~\ref{sec:initial_2D_heatmap}).
From the heatmaps, we extract 2D joint positions as anchors on the heatmap features.
These features interact with joint queries 
in our refinement module to generate multi-view-aware offset features (Sec.~\ref{sec:heatmap_refinement}).
To better capture the state of the initial heatmap estimation, we enhance the joint queries using initial heatmaps and RGB inputs (Sec.~\ref{sec:query}).
The offset features are then added to the initial heatmap features to obtain refined features and heatmaps, which can be used with existing 2D-to-3D lifting modules (Sec.~\ref{sec:3d_pose}).
Moreover, since initial 2D joint detection is not always accurate, we also utilize heatmap uncertainty to explicitly guide the refinement module to prioritize the heatmap features with higher confidence (Sec.~\ref{sec:uncertainty_aware_masking}).

Furthermore, we introduce two new large-scale datasets: Ego4View-Syn, \ie synthetic data based on the concept of our new HMD design, and Ego4View-RW, \ie real-world data captured with our new HMD prototype (Sec.~\ref{sec:dataset}).

In short, the contributions of this paper are as follows:

\begin{itemize} 
    \setlength{\itemsep}{1pt} 
    \item The first approach to study egocentric rear cameras, inspiring new HMD designs tailored for full-body tracking; 
    \item The simple and effective transformer-based module that refines 2D joint heatmap estimation with multi-view context and uncertainty-aware masking, which in turn boosts the accuracy of the estimated 3D human poses;  
    \item New benchmark datasets, \textit{Ego4View-Syn} and \textit{Ego4View-RW} for egocentric 3D full-body tracking with rear views. 
\end{itemize}

Our experiments show that the new configurations with rear cameras can substantially improve 3D body tracking while offering other sensing benefits, as mentioned before. 
Furthermore, the proposed method addresses the challenges associated with the integration of rear views (\ie unreliable 2D joint detection), achieving further improvements over the current state of the art, \eg ${>}10\%$ on Ego4View-RW.
We release the source code, trained models, Ego4View-Syn, and Ego4View-RW for the egocentric vision community.

%% file: sec/2_related_work.tex
\section{Related Work}
\label{sec:related_work}
\noindent\textbf{Egocentric Human Motion Capture.} 
Recent years have seen substantial progress in egocentric human motion capture, with existing works utilizing body-facing cameras to capture HMD users, broadly classified into two groups.
The first group includes monocular methods~\cite{xu2019mo2cap2, tome2019xr, Tom2020SelfPose3E, zhang2021automatic, wang2021estimating, park2023domain, wang2022estimating, jiang2021egocentric, yuan2019ego, luo2021dynamics, Liu2023, Millerdurai_EventEgo3D_2024, wang2024egocentric}, which struggle to accurately estimate 3D poses due to depth ambiguities.
The second group, including our work, focuses on the multi-view methods. 
Some works~\cite{zhao2021egoglass, hakada2022unrealego} use multi-branch auto-encoder while
Kang~\etal~\cite{kang2023ego3dpose, kang2024egotap} integrate optical feature extraction and stereo matching, and present a pose lifting module with an attention mechanism.
Akada~\etal~\cite{hakada2024unrealego2} utilize temporal and scene information.
EgoPoseFormer~\cite{yang2024egoposeformer} is the current state-of-the-art method utilizing deformable attention~\cite{zhu2021deformable} to update 3D poses.
However, existing methods easily fail to estimate accurate 3D poses in certain scenarios with self-occlusion or missing body parts in egocentric (back) views, due to their dependence on individual 2D joint detectors without effectively utilizing multi-view context.

To address this issue, we propose a simple but effective transformer-based module that refines 2D joint heatmap estimation with multi-view information and heatmap uncertainty.
The proposed module can be integrated with existing 2D joint heatmap estimators and 2D-to-3D lifting modules.

\begin{figure*}[t]
 \centering
 \includegraphics[width=\linewidth]{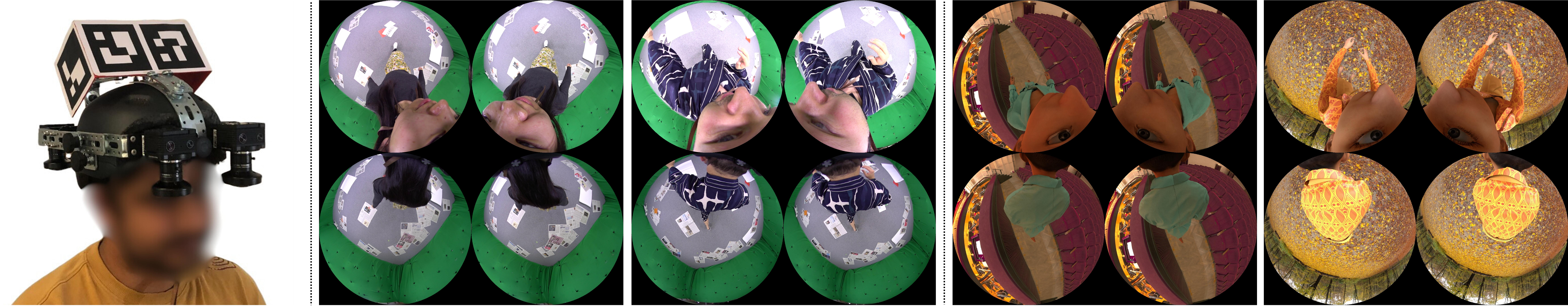}
    \vspace{-7mm}
    \caption*{\footnotesize{
        Our prototypical HMD
        \,\,\,\,\,\,\,\,\,\,\,\,\,\,\,\,\,\,\,\,\,\,\,\,\,\,
        Egocentric images sampled from Ego4View-RW
        \,\,\,\,\,\,\,\,\,\,\,\,\,\,\,\,\,\,\,\,\,\,\,\,\,\,
        Egocentric images sampled from Ego4View-Syn
        \,\,\,
    }}
 \vspace{-2mm}
 \caption{\textbf{Our new setup (Left) and egocentric images sampled from Ego4View-RW (Middle) and Ego4View-RW (Right). 
 } 
 }
 \label{fig:device} 
 \vspace{-3mm}
\end{figure*}

\begin{table}[t]
\vspace{-1mm}
\centering
\scalebox{0.705}{
\begin{tabular}{lccccccccc}
\hline
 Synthetic   & V & Id & Img & GT & Def & LC & Ha  & SMPL & RC \\
\hline
  M2C2~\cite{xu2019mo2cap2} & 1 & - & 530k & 530k & \xmark & \xmark & \xmark & \xmark & \xmark \\
 xR-EP~\cite{tome2019xr} & 1 & 34 & 380k & 380k & \xmark & \xmark & \cmark & \xmark & \xmark \\
 EgoWB~\cite{wang2024egocentric}  & 1  & 14 & 700k & 700k & \xmark & \xmark & \cmark & \xmark & \xmark \\
 UE~\cite{hakada2022unrealego} & 2 & 17 & 900k & 450k & \xmark & \xmark & \cmark & \xmark & \xmark \\
 UE2~\cite{hakada2024unrealego2}  & 2 & 17 & 2.5M & 1.2M & \xmark & \xmark & \cmark & \xmark & \xmark \\
 SE~\cite{cuevas2024simpleego}  & 1 & 6k & 120k & 60k & \xmark & \xmark & \cmark & \cmark & \xmark \\
\cdashline{1-10}
\noalign{\smallskip}
 \bf Ego4View-Syn & 4 & 111 & 1.5M & 375k & \cmark & \cmark & \cmark & \cmark & \cmark \\
\noalign{\smallskip}
\hline
\hline
 Real-World   & V & Id & Img & GT & Def & LC & Ha  & SMPL & RC \\
\hline
xR-EP-R~\cite{tome2019xr}* & 1 & - & 10k & 10k & \cmark & \xmark & \xmark & \xmark & \xmark \\
  EgoGlass~\cite{zhao2021egoglass}*  & 2 & 10 & 173k & 86k & \cmark & \xmark & \xmark & \xmark & \xmark \\
 EPW~\cite{wang2022estimating} $^\star$  & 1 & 10 & 318k & 318k & \cmark & \xmark & \xmark & \xmark & \xmark \\
 M2C2-R~\cite{xu2019mo2cap2} & 1 & 3 & 5k & 5k & \cmark & \xmark & \xmark & \xmark & \xmark \\
 EgoGlobal~\cite{wang2021estimating}  & 1 & 2 & 12k & 12k & \cmark & \xmark & \xmark & \xmark & \xmark \\
 SceneEgo~\cite{wang2023scene}  & 1 & 5 & 82k & 82k & \cmark & \xmark & \xmark & \xmark & \xmark \\
 EgoCap~\cite{rhodin2016egocap}  & 2 & 8 & 75k & 37k & \cmark & \xmark & \xmark & \xmark & \xmark \\ 
 UE-RW~\cite{hakada2024unrealego2} & 2 & 16 & 260k & 130k & \cmark & \xmark & \xmark & \xmark & \xmark \\
\cdashline{1-10}
\noalign{\smallskip}
 \bf Ego4View-RW & 4 & 25 & 930k & 232k & \cmark & \cmark & \cmark & \cmark & \cmark \\
\noalign{\smallskip}
\hline
\end{tabular}
}
\vspace{-2mm}
\caption{\textbf{Comparison of existing datasets for egocentric 3D human pose estimation using body-facing RGB cameras.} 
``\textbf{$\ast$}'': not publicly available; ``\textbf{$\star$}'': pseudo ground truths; 
\textbf{V}: the number of views; \textbf{Id}: the number of human identities; \textbf{Img}: the number of images; \textbf{GT}: the number of ground-truth 3D poses; \textbf{Def}: realistic cloth deformation; \textbf{LC}: loose clothes; \textbf{Ha}: hand annotations; \textbf{SMPL}: SMPL parameters \cite{SMPL:2015}; \textbf{RC}: rear cameras.
}
\label{table:comparison_dataset}
\vspace{-2mm}
\end{table}

\begin{table}[h]
\centering
\scalebox{0.75}{
\begin{tabular}{l||l|cc:cc}
\hline
D-FH & View  & Hand-L & Hand-R  & Foot-L & Foot-R  \\
\hline
\multirow{2}{*}{$5\sim7$\text{cm}} & Front-L  & 61.8\% & 52.4\%  & 11.1\% & 12.8\% \\
& Front-R  & 47.6\% & 65.8\%  & 12.8\% & 11.2\% \\
\hline
\hline
\begin{tabular}[l]{@{}c@{}}D-RH (D-FR) \end{tabular} & View  & Hand-L & Hand-R  & Foot-L & Foot-R  \\
\hline
\multirow{2}{*}{\begin{tabular}[l]{@{}c@{}}$3\sim5$\text{cm}  (32\text{cm}) \end{tabular}} & Rear-L  & 25.0\% & 8.3\%  & 6.1\% & 4.6\% \\
& Rear-R  & 10.3\% & 27.1\%  & 4.9\%  & 5.3\%  \\
\cdashline{1-6}
\multirow{2}{*}{\begin{tabular}[l]{@{}c@{}}$8\sim10$\text{cm}  (37\text{cm}) \end{tabular}} & Rear-L  & 25.2\% & 14.8\%  & 14.6\%  & 12.5\%  \\
& Rear-R  & 16.2\% & 25.1\%  & 13.3\%  & 13.1\%  \\
\cdashline{1-6}
\multirow{2}{*}{\begin{tabular}[l]{@{}c@{}}$13\sim15$\text{cm}  (42\text{cm}) \end{tabular}} & Rear-L  & 22.9\% & 12.7\%  & 30.5\% & 30.2\% \\
& Rear-R  & 15.5\% & 22.1\%  & 30.3\%  & 29.7\%  \\
\hline
\end{tabular}
}
\vspace{-2mm}
\caption{
\textbf{Visibility analysis of end-effector joints (hands and feet) with various rear-camera settings in \%, \ie~the percentage of cases with no occlusion.} 
\textbf{D-FR}: distance from front to rear cameras.
\textbf{D-FH}: expected distance from front cameras to the front of the human head.
\textbf{D-RH}: expected distance from rear cameras to the back of the human head.
\textbf{-L}: left. 
\textbf{-R}: right.
}
\label{table:joint_visibility}
\vspace{-5mm}
\end{table}

\noindent\textbf{Transformers in Outside-in 3D Human Pose Estimation.} 
Recent advances of Transformer~\cite{vaswani2017attention} have significantly improved 3D full-body tracking using external cameras.
Many approaches~\cite{pavllo:videopose3d:2019, zhu2021posegtac, Zheng_2021_ICCV, Li_2022_CVPR, li2022exploiting, zhang2022mixste, yang2022u, einfalt2023uplift, Zhao_2023_CVPR, Tang_2023_CVPR, Park_2023_ICCV, Zhu_2023_ICCV, Zhou_2023_ICCV} propose monocular 3D lifting modules to estimate 3D poses from 2D joint inputs. 
While these methods yield impressive results, they are not readily adaptable to multi-view settings.
Alternatively, several studies~\cite{He_2020_CVPR, ma2021transfusion} utilize stereo epipolar geometry. 
However, this is challenging to achieve with fisheye images due to the significant distortion.

In this work, we leverage initial joint heatmaps to obtain 2D joint positions, serving as anchors on the corresponding heatmap features, and facilitating the extraction of multi-view features to refine the heatmap estimation.

\noindent\textbf{HMD Setups.} 
In egocentric tasks, \eg 
hand tracking~\cite{li2013model, bambach2015lending, liu2024single, EgoPressure, Grauman_etal_2022_CVPR, Grauman_2024_CVPR}, it is standard practice to use cameras positioned in front of the human body, often with HMDs.
This setup is similarly prevalent in egocentric human pose estimation, where existing works utilize frontal cameras with slight variations in positioning, such as forward-facing~\cite{jiang2021egocentric, yuan2019ego, luo2021dynamics, Zhang_ECCV_2022, Pan_2023_ICCV, Khirodkar_2023_ICCV, Zhang_2023_ICCV, Li2023EgoBody, Grauman_etal_2022_CVPR, Pan_2023_ICCV, Khirodkar_2023_ICCV, Grauman_2024_CVPR, hollidtegosim, chi2025estimating} or body-facing cameras~\cite{rhodin2016egocap, xu2019mo2cap2, tome2019xr, wang2021estimating, wang2022estimating, jiang2021egocentric, yuan2019ego, luo2021dynamics, zhao2021egoglass, Li2023EgoBody}.
However, this camera placement may not be optimal for egocentric full-body motion capture as discussed in Sec.~\ref{sec:intro}.
Specifically, the back of the body is not captured at all, despite its potential value in providing crucial 3D reconstruction cues.

To explore this setup, we develop a new HMD prototype using front and rear cameras, detailed in the next section. 
\textbf{Note that this is the first work to investigate the potential of body-facing rear cameras installed on the HMD.}

%% file: sec/3_dataset.tex
\section{Setup and Dataset}
\label{sec:dataset}

\noindent\textbf{Design of Our HMD Prototype.}
To explore HMD designs tailored for full-body tracking, we build a new HMD prototype (Fig.~\ref{fig:device}). 
This setup is based on a helmet with 4 downward-facing RIBCAGE RX0 \text{I\hspace{-1.2pt}I} cameras~\cite{ribcagecamera} (two on the front and two on the back). 
Each camera comes with a FUJINON FE185C057HA-1 fisheye lens~\cite{fujinonlens}.
Right and left cameras are placed 12 cm apart, while the front and rear cameras are installed 37 cm away from each other.

\noindent\textbf{Ego4View-RW (Real-World) Dataset.}
We record motions of 25 identities in a motion capture studio with 120 cameras.
HMD users wear not only simple tight clothes typically seen in existing data~\cite{hakada2024unrealego2, wang2021estimating, rhodin2016egocap, wang2023scene} but also loose ones (long skirts, yukata, etc.) that bring new challenges 
in egocentric tasks. 
We follow UnrealEgo-RW~\cite{hakada2024unrealego2} to capture both simple and complex motions (walking, crawling, dancing, etc.). 
Overall, Ego4View-RW offers 930,812 images
of a resolution $872{\times}872$ pixel captured at $25$ frames per second for 2.6 hours.
We obtain SMPL parameters using multi-view video streams and EasyMocap~\cite{easymocap}.
Note that Ego4View-RW offers $1.8 \times$ larger number of 3D poses than the existing real-world data~\cite{hakada2024unrealego2} (Table~\ref{table:comparison_dataset}).

\begin{figure*}[t]
 \centering
  \includegraphics[width=\linewidth]{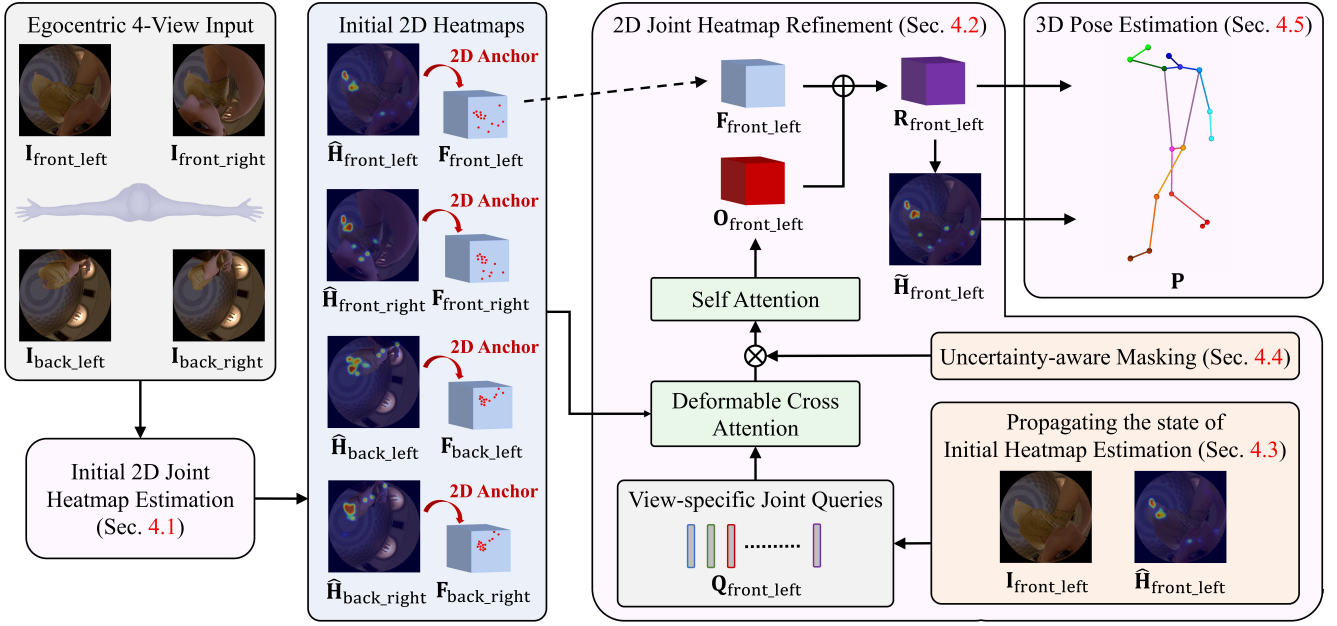}
 \vspace{-6mm}
 \caption{
 \textbf{Overview of our framework.} 
 Given front and rear views $\mathbf{I}$, we first use a 2D joint estimator to obtain 2D joint heatmaps $\widehat{\mathbf{H}}$ (Sec.~\ref{sec:initial_2D_heatmap}). 
 We also extract 2D joint positions from the heatmaps as anchors on the corresponding heatmap feature $\mathbf{F}$. 
 Next, our refinement module refines the heatmap estimation for each view (Sec.~\ref{sec:heatmap_refinement}). 
 Here, we show an example process to refine the heatmap for the front-left view.
 Our view-specific joint queries $\mathbf{Q}_{\text{front\_left}}$ interact with the features $\mathbf{F}$ of all views around the anchors in our refinement module to generate multi-view-aware offset features $\mathbf{O}_{\text{front\_left}}$. 
 To better capture the initial heatmap estimation state, we enhance the joint queries with embeddings of the initial heatmap $\widehat{\mathbf{H}}_{\text{front\_left}}$ and the RGB input $\mathbf{I}_{\text{front\_left}}$ (Sec.~\ref{sec:query}).
 Furthermore, we utilize heatmap uncertainty to explicitly guide the refinement module to prioritize the heatmap features with higher confidence (Sec.~\ref{sec:uncertainty_aware_masking}).
 The offset features $\mathbf{O}_{\text{front\_left}}$ are added to initial heatmap features $\mathbf{F}_{\text{front\_left}}$ to obtain refined features $\mathbf{R}_{\text{front\_left}}$ and heatmaps $\widetilde{\mathbf{H}}_{\text{front\_left}}$.
 We iterate this refinement process for all views and resulting refined features $\mathbf{R}_{k}$ and heatmaps $\widetilde{\mathbf{H}}_{k}$ can be used with existing 2D-to-3D lifting modules to estimate a 3D pose $\mathbf{P}$ (Sec.~\ref{sec:3d_pose}).
 } 
  \vspace{-2mm}
 \label{fig:overview_method} 
\end{figure*}

\noindent\textbf{Ego4View-Syn (Synthetic) Dataset.}
Based on our HMD setup, we create a large-scale synthetic dataset using Blender.
Existing synthetic data~\cite{xu2019mo2cap2, tome2019xr, hakada2024unrealego2, Millerdurai_EventEgo3D_2024, wang2024egocentric} lack realistic cloth deformation, making it impossible to evaluate methods in realistic scenarios.
To solve this issue, we animate 111 SMPL models~\cite{SMPL:2015} using clothing and motion data from BEDLAM~\cite{Black_CVPR_2023} (with all of their motion types).
In total, Ego4View-Syn offers 1,499,692 images with a resolution $872{\times}872$ pixels.
Each frame is annotated with SMPL parameters, from which we use 16 joints in our work.
See Table~\ref{table:comparison_dataset} for detailed comparisons with existing datasets.

\noindent\textbf{Joint Visibility.} 
Table~\ref{table:joint_visibility} analyses the visibility of end-effector joints (hands and feet) with various rear-view settings in our synthetic setup.
This reveals that rear views exhibit more frequent occlusions than front views, particularly for hands, which often leads to mis-detection of 2D positions (discussed in Sec.~\ref{subsec:experimental_results} with Fig.~\ref{fig:qualitaitve_result_hm2d}-(bottom left)).
Moreover, while farther rear cameras improve foot visibility, they do not enhance hand visibility, likely due to hand movements behind the body as well as closer rear cameras occasionally capturing hands over shoulders.
Based on the analysis, we decided to use our prototypical setup with 37 cm as a default distance from front to rear cameras, considering the trade-off between visibility and the form factor.

%% file: sec/4_method.tex
\section{Method}
\label{sec:method}

As discussed in Sec.~\ref{sec:intro}, simply adding rear views to frontal inputs is not optimal for existing methods or can sometimes degrade accuracy compared to using only frontal inputs. 
To address this issue, we propose a new transformer-based method that leverages multi-view information to refine 2D joint heatmaps, thereby improving the accuracy of the estimated 3D poses.
The proposed module is motivated by the implicit assumption that unreliable rear-view heatmaps can be improved by reliable front-view heatmaps and vice versa, due to the symmetrical nature of the human body.
See Fig.~\ref{fig:overview_method} for an overview of our approach.
We first regress 2D joint heatmaps from egocentric images (Sec.~\ref{sec:initial_2D_heatmap}). 
From the heatmaps, we extract 2D joint positions that serve as anchors for the corresponding heatmap features. 
Subsequently, the heatmap features from all views engage with view-specific joint queries via attention mechanisms in our refinement module (Sec.~\ref{sec:heatmap_refinement}). 
This interaction generates offset features representing the discrepancies between the initial heatmaps and the ground truth. 
To further improve the representation of the initial heatmap estimations, we enhance the view-specific joint queries with embeddings derived from initial heatmaps and fisheye RGB images (Sec.~\ref{sec:query}). 
Moreover, since the initial 2D joint detection is not always accurate (due to self-occlusion or missing body parts), we incorporate uncertainties from the initial heatmaps (Sec.~\ref{sec:uncertainty_aware_masking}), guiding the refinement module to prioritize heatmap features with higher confidence. 
The resultant offset features are added to initial heatmap features to yield refined features and heatmaps, which can be used with existing 2D-to-3D lifting modules (Sec.~\ref{sec:3d_pose}). 

\subsection{Initial 2D Joint Heatmap Estimation}
\label{sec:initial_2D_heatmap}

Given front and rear view inputs
$\mathbf{I} = \{\mathbf{I}_{k} \in \mathbb{R}^{H \times W \times 3 } \}$ where $k = \{ \text{front\_left}, \text{front\_right}, \text{back\_left}, \text{back\_right} \}$, 
we use an existing 2D joint heatmap estimator~\cite{yang2024egoposeformer, kang2024egotap} to obtain 2D heatmaps of 15 joints 
$\widehat{\mathbf{H}} =  \{\widehat{\mathbf{H}}_{k} \in \mathbb{R}^{\frac{H}{4} \times \frac{W}{4} \times 15 }\}$, 
including the neck, arms, forearms, hands, upper legs, legs, feet, and toes, as well as down-sampled heatmap features from the final regression layer 
$ \mathbf{F} = \{\mathbf{F}_{k} \in \mathbb{R}^{\frac{H}{8} \times \frac{W}{8} \times 256 }\}$.
We also extract 2D joint positions by identifying the maximum values of the heatmaps, which will serve as anchors for the heatmap features in our refinement module.

\subsection{2D Joint Heatmap Refinement}
\label{sec:heatmap_refinement}

We utilize multi-view context to refine initial heatmap features for each view.
Specifically, our objective is to generate a refinement offset relative to the initial heatmap features for each view.
For clarity, we detail the refinement process applied to the heatmap of the front-left view $\widehat{\mathbf{H}}_{\text{front\_left}}$.

In our refinement module, we employ a transformer-decoder-based architecture with \textit{joint queries}.
The existing works~\cite{hakada2024unrealego2, yang2024egoposeformer} use joint queries to represent 3D body joints as learnable embeddings, encoding prior knowledge of skeletal structures within a device-relative 3D space.
In contrast, our refinement module utilizes view-specific joint queries $\mathbf{Q}_{\text{front\_left}} \in \mathbb{R}^{15 \times 256}$ to embed skeletal information within a 2D space specific to the egocentric front-left view.

The view-specific joint queries facilitate the extraction of joint-level multi-view information from the heatmap features across all views in attention mechanisms.
Note that since the initial heatmaps already highlight coarse regions for potential 2D joint positions, applying attention to a focused subset of heatmap features around the 2D anchors---rather than across the full spatial dimensions---enhances efficiency without compromising precision.
Specifically, we adopt deformable attention~\cite{zhu2021deformable}, enabling the interaction of the view-specific joint queries with a subset of heatmap features of all $k$ views around the 2D anchors $\textbf{T}_{k} \in \mathbb{R}^{ 15 \times 2 }$:
\begin{equation}
  \widehat{\mathbf{Q}}^{k}_{\text{front\_left}} = \text{DeformAttn}(\mathbf{Q}_{\text{front\_left}}, \textbf{T}_{k}, \mathbf{F}_{k}). 
    \label{eq:deform_attn}
\end{equation}
The updated queries $\widehat{\mathbf{Q}}^{k}_{\text{front\_left}}$ are concatenated along with the dimension $k$ and passed to a fully connected layer to obtain multi-view-aware joint query for the front-left view $\widehat{\mathbf{Q}}_{\text{front\_left}} \in \mathbb{R}^{15 \times 256}$. 
Next, these updated queries interact with each other via standard self-attention:
\begin{equation}
  \widetilde{\mathbf{Q}}_{\text{front\_left}} = \text{SelfAttn}(\widehat{\mathbf{Q}}_{\text{front\_left}}).
  \label{eq:self_attn}
\end{equation}
Now, the queries $\widetilde{\mathbf{Q}}_{\text{front\_left}}$ contain multi-view-aware 2D-joint knowledge tailored for the front-left view.
Thus, we utilize these queries to generate offset features that capture discrepancies between initial heatmaps and ground truths for the front-left view.
Specifically, we feed the reshaped queries $\widetilde{\mathbf{Q}}_{\text{front\_left}} \in \mathbb{R}^{15 \times 16 \times 16}$ into an offset regression network $\mathcal{F}_{\text{O}}(\cdot)$ 
to obtain multi-view-aware offset features $\mathbf{O}_{\text{front\_left}} \in \mathbb{R}^{128 \times 32 \times 32}$:
\begin{equation}
  \mathbf{O}_{\text{front\_left}} = \mathcal{F}_{\text{O}}(\widetilde{\mathbf{Q}}_{\text{front\_left}}).
\end{equation}
The offset feature is then added to the initial heatmap features for the left view $\mathbf{F}_{\text{front\_left}}$ to generate refined heatmap features $\mathbf{R}_{\text{front\_left}} \in \mathbb{R}^{128 \times 32 \times 32}$:
\begin{equation}
  \mathbf{R}_{\text{front\_left}} = \mathcal{F}_{\text{R}}(\mathbf{F}_{\text{front\_left}} + \mathbf{O}_{\text{front\_left}}),
\end{equation}
where ``$+$'' denotes element-wise addition and $\mathcal{F}_{\text{R}}$ is an offset projection layer.
Finally, the refined features are fed into a heatmap regression layer $\mathcal{F}_{\text{HM}}(\cdot)$:
\begin{equation}
  \widetilde{\mathbf{H}}_{\text{front\_left}} = \mathcal{F}_{\text{HM}}(\mathbf{R}_{\text{front\_left}}).
\end{equation}
The final heatmaps $\widetilde{\mathbf{H}}_{\text{front\_left}}$ and features $\mathbf{R}_{\text{front\_left}}$ can be used with existing 3D modules to infer a 3D pose (Sec.~\ref{sec:3d_pose}). 
Note that the process described so far is repeated for all $k$ views, \ie$\{ \text{front\_left}, \text{front\_right}, \text{back\_left}, \text{back\_right}\}$.

Our refinement module is supervised with the MSE loss:
\begin{equation}
\label{eq:losshms}
\mathcal{L}_{\text{Refine}} = \frac{1}{J}\sum_{j=1}^{J} \lVert \mathbf{\widehat H}^{j}_{k} - \mathbf{H}^{j}_{k} \rVert ^2, 
\end{equation}
where $J$ is the number of body joints; $\mathbf{\widehat H}^{j}_{k}$ and $\mathbf{H}^{j}_{k}$ are the predicted and ground-truth heatmaps of the $j$-th joint.

\subsection{Propagating Initial Heatmap Estimation State}
\label{sec:query}
The refinement process (Sec.~\ref{sec:heatmap_refinement}) aims to derive refinement offsets by processing view-specific joint queries with multi-view heatmap features.
However, directly integrating these queries into attention mechanisms can be suboptimal for the offset feature generation, as they only capture the 2D skeletal information specific to each view, while lacking context from the initial heatmap predictions specific to each view. 

Hence, we embed the initial heatmaps and corresponding RGB images into the queries for each view, thereby propagating the initial heatmap estimation state more effectively. 
Firstly, we feed the initially estimated heatmaps $\widehat{\mathbf{H}}_{\text{front\_left}}$ into an MLP projection layer $\mathcal{P}_{\text{HM}}(\cdot)$ to generate the heatmap embeddings $\mathbf{E}_{\text{front\_left}} \in \mathbb{R}^{15 \times 256}$:
\begin{equation}
  \mathbf{E}_{\text{front\_left}} = \mathcal{P}_{\text{HM}}(\widehat{\mathbf{H}}_{\text{front\_left}}).
\end{equation}
We also process RGB features $\mathbf{B}_{\text{front\_left}} \in \mathbb{R}^{512 \times 8 \times 8}$ taken from the encoder backbone of the 2D joint estimator via an MLP projection layer $\mathcal{P}_{\text{RGB}}(\cdot)$ to generate RGB embeddings $\mathbf{G}_{\text{front\_left}} \in \mathbb{R}^{256 \times 1 \times 1}$:
\begin{equation}
  \mathbf{G}_{\text{front\_left}} = \mathcal{P}_{\text{RGB}}(\mathbf{B}_{\text{front\_left}}).
\end{equation}
These embeddings are added to the joint queries and passed into a query projection layer $\mathcal{P}_{\text{Q}}(\cdot)$:
\begin{equation}
  \mathbf{Q'}_{\text{front\_left}} = \mathcal{P}_{\text{Q}}(\mathbf{Q}_{\text{front\_left}} + \mathbf{E}_{\text{front\_left}} +  \mathbf{G}_{\text{front\_left}}),
\end{equation}
where $\mathbf{Q'}_{\text{front\_left}}$ represents updated joint queries that encapsulate information regarding the state of the initial heatmap estimation.
Finally, we replace the original queries $\mathbf{Q}_{\text{front\_left}}$ in Eq.~\eqref{eq:deform_attn} with the new queries $\mathbf{Q'}_{\text{front\_left}}$.

\subsection{Uncertainty-aware Masking}
\label{sec:uncertainty_aware_masking}
Egocentric images often exhibit substantial self-occlusions and missing body part observations. 
This leads to unreliable initial 2D joint heatmap estimation, producing inaccurate 2D anchor points in certain views.
To mitigate this issue, we incorporate heatmap uncertainty, directing the network to prioritize heatmaps with higher confidence.
Specifically, we construct uncertainty-based masks $\mathbf{M}^{k} \in \mathbb{R}^{15 \times 1} $ by assessing the heatmap values at the 2D joint positions:

\begin{equation}
    \mathbf{M}^{k} = 
        \begin{cases}
            1, & \text{if heatmap values ${\geq}$ 0.5} \\
            0, & \text{otherwise}
        \end{cases}.
\end{equation}
These masks are applied as element-wise multipliers to the updated queries $\widehat{\mathbf{Q}}^{k}_{\text{front\_left}}$ in Eq.~\eqref{eq:deform_attn}:

\begin{equation} \widehat{\mathbf{Q'}}^{k}_{\text{front\_left}} = \widehat{\mathbf{Q}}^{k}_{\text{front\_left}} \times \mathbf{M}^{k},
\end{equation}
where ``$\times$'' denotes element-wise multiplication. 
Finally, we replace the queries $\widehat{\mathbf{Q}}^{k}_{\text{front\_left}}$ in Eq.~\eqref{eq:deform_attn} with the newly constructed one $\widehat{\mathbf{Q'}}^{k}_{\text{front\_left}}$.
This enables the subsequent self-attention operation in Eq.~\eqref{eq:self_attn} to direct greater focus toward heatmap values with higher confidence.

\subsection{3D Human Pose Estimation}
\label{sec:3d_pose}
The refined heatmap features  $\mathbf{R}_{k}$ and heatmaps $\widetilde{\mathbf{H}}_{k}$ can be utilized with existing 2D-to-3D lifting modules to estimate 3D poses.
Thus, we integrate our module into the current state-of-the-art methods (Sec.~\ref{subsec:experimental_results}), EgoPoseFormer~\cite{yang2024egoposeformer} and EgoTAP~\cite{kang2024egotap}, to see the efficacy of our method.

%% file: sec/5_experiment.tex
% ----------------------------------------------------------------- %
\begin{table}[t]
\centering
\scalebox{0.73}{
\begin{tabular}{clcccc}
\hline
\noalign{\smallskip}
& & \multicolumn{2}{c}{Ego4View-Syn} & \multicolumn{2}{c}{Ego4View-RW} \\
\noalign{\smallskip}
\cdashline{3-6}
\noalign{\smallskip}
\multicolumn{1}{c}{Setup} & \multicolumn{1}{c}{Method} &  \multicolumn{1}{c}{MPJPE} & \multicolumn{1}{c}{\begin{tabular}[c]{@{}c@{}}PA- \\MPJPE \end{tabular}} & \multicolumn{1}{c}{MPJPE} & \multicolumn{1}{c}{\begin{tabular}[c]{@{}c@{}}PA- \\MPJPE \end{tabular}} \\ 
\noalign{\smallskip}
\hline
\noalign{\smallskip}
\multirow{1}{*}{\begin{tabular}[c]{@{}c@{}}2 rear views \end{tabular}} & 
EPFormer~\cite{yang2024egoposeformer} &  52.90 & 43.99 & 114.95 & 100.11 \\
\noalign{\smallskip}
\hline
\noalign{\smallskip}
\multirow{4}{*}{\begin{tabular}[c]{@{}c@{}}2 front views \\ (no rear views) \end{tabular}} & 
EgoTAP~\cite{kang2024egotap} &  32.56 & 27.44 & 91.23 & 79.10 \\
& \,\, + Ours & \bf 31.90 & \bf 26.85  & \bf 89.14 & \bf 77.28 \\
\noalign{\smallskip}
\cdashline{2-6}
\noalign{\smallskip}
& EPFormer~\cite{yang2024egoposeformer} &  27.36 & 23.31 & 77.95 & \bf 67.49 \\
& \,\, + Ours & \bf 27.04  & \bf 23.18 & \bf 76.35  & 67.97  \\
\noalign{\smallskip}
\hline
\noalign{\smallskip}
\multirow{4}{*}{\begin{tabular}[c]{@{}c@{}}2 front views \\ + \\ 1 rear-left view \end{tabular}} & EgoTAP~\cite{kang2024egotap} &  24.90 & 21.33 & 72.25 & 65.36 \\
& \,\, + Ours & \bf 24.12 & \bf 20.63 & \bf 67.09 & \bf 61.27 \\
\noalign{\smallskip}
\cdashline{2-6}
\noalign{\smallskip}
& EPFormer~\cite{yang2024egoposeformer}  & 21.51 & 18.78 & 66.08 & 59.90 \\
& \,\, + Ours & \bf 21.04 & \bf 18.60 & \bf 60.96 & \bf 55.83 \\
\noalign{\smallskip}
\hline
\noalign{\smallskip}
\multirow{4}{*}{\begin{tabular}[c]{@{}c@{}}2 front views \\ + \\ 1 rear-right view \end{tabular}} & EgoTAP~\cite{kang2024egotap}  &  25.08 & 21.41 & 72.41 & 65.74 \\
& \,\, + Ours & \bf 24.29 & \bf 20.73 & \bf 66.96 & \bf 61.08 \\
\noalign{\smallskip}
\cdashline{2-6}
\noalign{\smallskip}
& EPFormer~\cite{yang2024egoposeformer}  & 21.79 & 19.02 & 66.45 & 60.57 \\
& \,\, + Ours  & \bf 21.50 & \bf 18.97 & \bf 60.17 & \bf 54.92 \\
\noalign{\smallskip}
\hline
\noalign{\smallskip}
\multirow{4}{*}{\begin{tabular}[c]{@{}c@{}}2 front views \\ + \\ 2 rear views \end{tabular}} & EgoTAP~\cite{kang2024egotap}  &  23.88 & 20.93 & 69.78 & 64.19 \\
& \,\, + Ours & \bf 22.57 & \bf 19.80 & \bf 62.11 & \bf 57.13 \\
\noalign{\smallskip}
\cdashline{2-6}
\noalign{\smallskip}
& EPFormer~\cite{yang2024egoposeformer} & 20.20 & 17.63 & 63.38 & 58.25 \\
& \,\, + Ours  & \bf 19.25 & \bf 16.95 & \bf 56.94 & \bf 52.25 \\
\noalign{\smallskip}
\hline
\end{tabular}
}
\vspace{-3mm}
\caption{\textbf{Quantitative results on Ego4View-Syn and Ego4View-RW} with mm-scale. \textbf{EPFormer}: EgoPoseFormer~\cite{yang2024egoposeformer}. We also provide EgoPoseFormer~\cite{yang2024egoposeformer} with 2 rear views as a reference.
}
\label{table:quantitative_result}
\vspace{-2mm}
\end{table}

\section{Experiments} 
\label{sec:experiments}

\subsection{Datasets and Training Details}
\label{subsec:training_details}

\noindent\textbf{Datasets.} 
We use the proposed datasets for rear-view evaluation. 
We split Ego4View-Syn into 5,020 motions (900,152 views) for training, 1,613 motions (288,148 views) for validation, and 1,739 motions (311,392 views) for testing. 
Similarly, we divide Ego4View-RW into 286 motions (557,484 views) for training, 102 motions (198,732 views) for validation, and 90 motions (174,596 views) for testing.

\begin{table}[t]
\centering
\scalebox{0.69}{
\begin{tabular}{lccccccc}

\noalign{\smallskip}
\hline
\noalign{\smallskip}
\multicolumn{1}{l}{Ego4View-Syn} & head & neck & arms & forearms & hands & \begin{tabular}[c]{@{}c@{}}upper \\ legs \end{tabular}  \\ 
\noalign{\smallskip}
\hline
\noalign{\smallskip}

EgoPoseFormer~\cite{yang2024egoposeformer}  & \bf 0.53  & \bf 3.87 & 8.08 & 13.54 & 22.05 & 15.30 \\

\,\, + Ours & \bf 0.53 & 4.17  & \bf 7.78 & \bf 12.94 & \bf 20.11 & \bf 14.79 & \\

\noalign{\smallskip}
\hline
\noalign{\smallskip}
\multicolumn{1}{l}{Ego4View-Syn} & legs & feet & toes & \begin{tabular}[c]{@{}c@{}} upper \\ body \end{tabular} & \begin{tabular}[c]{@{}c@{}} lower \\ body \end{tabular} &  \multicolumn{1}{c}{all} \\
\noalign{\smallskip}
\hline
\noalign{\smallskip}

EgoPoseFormer~\cite{yang2024egoposeformer}  & 26.87 & 35.45 & 38.07 & 11.48 & 28.92 & 20.20 \\

\,\, + Ours & \bf 25.75 & \bf 34.08 & \bf 36.22 & \bf 10.80 & \bf 27.71 & \bf 19.25 \\

\noalign{\smallskip}
\hline
\hline
\noalign{\smallskip}
\multicolumn{1}{l}{Ego4View-RW} & head & neck & arms & forearms & hands & \begin{tabular}[c]{@{}c@{}}upper \\ legs \end{tabular}  \\ 

\noalign{\smallskip}
\hline
\noalign{\smallskip}

EgoPoseFormer~\cite{yang2024egoposeformer}  
& 11.80
& 16.36
& 21.55
& 34.30
& 60.35
& 44.33 \\

\,\, + Ours  
& \bf 11.49
& \bf 15.89
& \bf 21.27
& \bf 30.90
& \bf 48.17
& \bf 42.32 \\

\noalign{\smallskip}
\hline
\noalign{\smallskip}
\multicolumn{1}{l}{Ego4View-RW} & legs & feet & toes & \begin{tabular}[c]{@{}c@{}} upper \\ body \end{tabular} & \begin{tabular}[c]{@{}c@{}} lower \\ body \end{tabular} &  \multicolumn{1}{c}{all} \\ 
\noalign{\smallskip}
\hline
\noalign{\smallskip}

EgoPoseFormer~\cite{yang2024egoposeformer}  

& 85.88
& 115.40
& 129.56
& 32.57
& 93.79
& 63.38 \\

\,\, + Ours  

& \bf 79.67
& \bf 103.46
& \bf 116.04
& \bf 28.51
& \bf 85.37
& \bf 56.94 \\

\noalign{\smallskip}
\hline
\noalign{\smallskip}
\end{tabular}
}
\vspace{-3mm}
\caption{\textbf{Per-joint evaluation} (MPJPE) with 2 front and 2 rear views.  \textbf{Top}: Ego4View-Syn. \textbf{Bottom}: Ego4View-RW.
}
\label{table:quantitative_result_joint}
\vspace{-5mm}
\end{table}

\noindent\textbf{Training Details.} 
We follow the existing works~\cite{hakada2022unrealego, hakada2024unrealego2, kang2023ego3dpose, kang2024egotap, yang2024egoposeformer} to use the input RGB images and ground-truth 2D joint heatmaps with $256{\times}256$ and $64{\times}64$ pixels, respectively. 
We use existing 2D joint heatmap estimators~\cite{yang2024egoposeformer, kang2024egotap} with its encoder pre-trained on ImageNet~\cite{imagenet_cvpr09}.
We first train the 2D joint heatmap estimator and the refinement module, respectively, for 12 epochs with AdamW~\cite{loshchilov2018decoupled}.
Next, we train the entire architecture, including the 3D modules taken from existing methods~\cite{yang2024egoposeformer, kang2024egotap} for 12 epochs.
The batch size is set to 64 for 2D heatmap estimation and 32 for 3D pose estimation.
The initial learning rate is set to $10^{-3}$ with a weight decay of $5 \cdot 10^{-3}$. 
The learning rate is decayed by a factor of 10 after the 8th and 10th epochs.

\begin{figure*}[t]
 \centering
 \includegraphics[width=\linewidth]{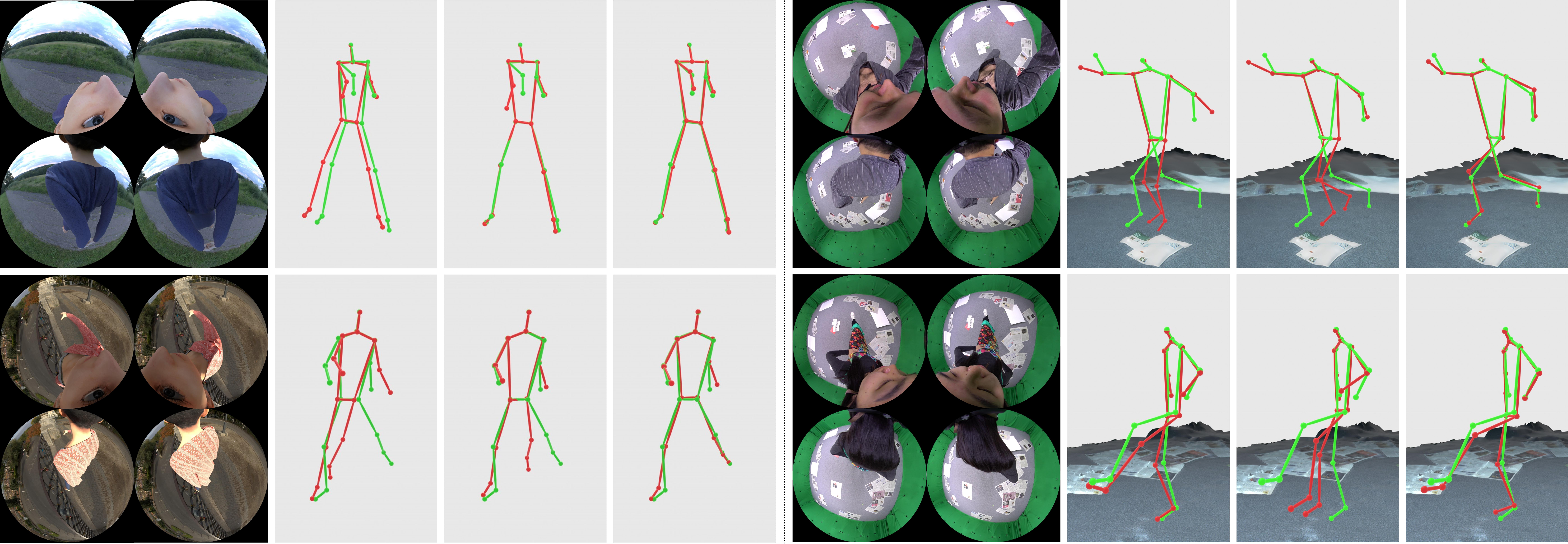}
    \vspace{-7mm}
    \caption*{\footnotesize{\,\,\,\,\,
        Multi-view inputs
        \,\,\,\,\,\,
        \begin{tabular}[l]{@{}c@{}} EgoPoseFormer \\ (2 front views)\end{tabular}
        \,
        \begin{tabular}[l]{@{}c@{}} EgoPoseFormer \\ (4 views) \end{tabular}
        \,\,
        \begin{tabular}[l]{@{}c@{}} Our method \\ (4 views)\end{tabular}
        \,\,\,\,\,\,\,\,\,\,\,\,\,\,\,\,\,\,
        Multi-view inputs
        \,\,\,\,\,\,\,
        \begin{tabular}[l]{@{}c@{}} EgoPoseFormer \\ (2 front views) \end{tabular}
        \,
        \begin{tabular}[l]{@{}c@{}} EgoPoseFormer \\ (4 views) \end{tabular}
        \,\,
        \begin{tabular}[l]{@{}c@{}} Our method \\ (4 views)\end{tabular}
    }}
    \vspace{-3mm}
 \caption{\textbf{Qualitative results of 3D pose estimation with EgoPoseFomer~\cite{yang2024egoposeformer} and our method.} 
 \textbf{Left}: Ego4View-Syn. \textbf{Right}: Ego4View-RW. 3D pose prediction and ground truth are displayed in red and green, respectively. 
 }
 \label{fig:qualitaitve_result_pose3d}
\vspace{-4.8mm}
\end{figure*}

\subsection{Experimental Evaluation} 
\label{subsec:experimental_results}
We compare our method with existing multi-view egocentric pose estimation methods~\cite{kang2024egotap, yang2024egoposeformer}.
EgoPoseFormer~\cite{yang2024egoposeformer} is the current state-of-the-art method, achieving 33 mm (MPJPE) on UnrealEgo~\cite{hakada2022unrealego}, whereas EgoTAP~\cite{hakada2022unrealego} is the second-best approach.
These methods are re-trained on our datasets with different camera configurations. 
For a fair comparison, we modify the source codes of the existing methods such that their 2D modules can take two front and two rear views as inputs. 
We report Mean Per Joint Position Error (MPJPE) and Mean Per Joint Position Error with Procrustes Alignment~\cite{10.1214/ss/1177012582} (PA-MPJPE).

\noindent\textbf{Impact of Rear Views.}
Table~\ref{table:quantitative_result} reports the quantitative results on Ego4View-Syn and Ego4View-RW.
Incorporating rear cameras enhances the average accuracy for all methods, \eg by 18.7\% on Ego4View-RW (MPJPE) with EgoPoseFormer~\cite{yang2024egoposeformer} from two front views (77.95 cm) to two front and two rear views (63.38 cm), even with more severe self-occlusions than frontal views.

Nonetheless, the closer examination of visual outputs (Fig.~\ref{fig:qualitaitve_result_pose3d}) reveals that EgoPoseFormer~\cite{yang2024egoposeformer} with four views often fails to accurately estimate 3D positions in many scenarios, \eg even when rear views clearly capture visible joints, such as the left arm (Fig.~\ref{fig:qualitaitve_result_pose3d}-(top left)) or the right leg (Fig.~\ref{fig:teaser} and Fig.~\ref{fig:qualitaitve_result_pose3d}-(top right)).
These results indicate that the current state-of-the-art method~\cite{yang2024egoposeformer} can not fully utilize the multi-view context.
Furthermore, Fig.~\ref{fig:qualitaitve_result_pose3d}-(bottom right) and the video (5:58$\sim$) on our project page show certain scenarios where EgoPoseFormer~\cite{yang2024egoposeformer} with four views generates lower-quality 3D poses than its two-view version.
This underscores our observation that although rear views help improve the average pose tracking accuracy, naively adding rear views to frontal inputs---even with the state-of-the-art method~\cite{yang2024egoposeformer}---is not optimal and thus, we need a dedicated method for 3D full-body tracking with rear cameras.

\noindent\textbf{Effect of Our Refinement Process.}
Our method addresses the above challenges associated with the rear-view integration.
Specifically, Table~\ref{table:quantitative_result} shows that our method outperforms the baseline method~\cite{kang2024egotap, yang2024egoposeformer} by a wide margin, \eg ${>}10\%$ on Ego4View-RW with two front and two rear views.
Note that our focus lies in making the best use of rear cameras, and our method is motivated by the implicit assumption that front and rear views could complement each other due to the symmetrical nature of the human body (Secs.~\ref{sec:intro} and ~\ref{sec:method}).
Thus, while our method achieves on-par or slightly better results in the two-front-view settings, \eg by $2.1$\% on Ego4View-RW (MPJPE) over EgoPoseFormer~\cite{yang2024egoposeformer}, our module is expected to be more effective with rear views, \ie ${>}10\%$ gains on Ego4View-RW.
Table~\ref{table:quantitative_result_joint} evaluates per-joint errors, where our refinement process constantly brings significant improvement, \eg by 20.2\% on hands and 10.3\% on feet on Ego4View-RW.
This indicates that the best-performing existing method~\cite{yang2024egoposeformer} struggles to accurately capture these critical end-effector joints even with four-view inputs, highlighting the importance of devising a method tailored for rear-view integration, like our approach.

Our qualitative evaluation (Fig.~\ref{fig:qualitaitve_result_pose3d}) suggests that the proposed method consistently delivers superior visual outputs compared to the baseline method~\cite{yang2024egoposeformer}.
Notably, our method is able to handle the challenging scenarios where the current state-of-the-art method~\cite{yang2024egoposeformer} with two front and two rear views under-performs their two-front-view version (Fig.~\ref{fig:qualitaitve_result_pose3d}-(bottom right) and video-(5:58$\sim$)).
These results validate the superiority of our approach, establishing it as a robust benchmark for egocentric pose tracking with rear-view integration and inspiring many future applications, including character animation (Fig.~\ref{fig:teaser}-(f)) as well as the development of new HMDs specifically tailored for full-body tracking.

\begin{figure*}[t]
 \centering
 \includegraphics[width=\linewidth]{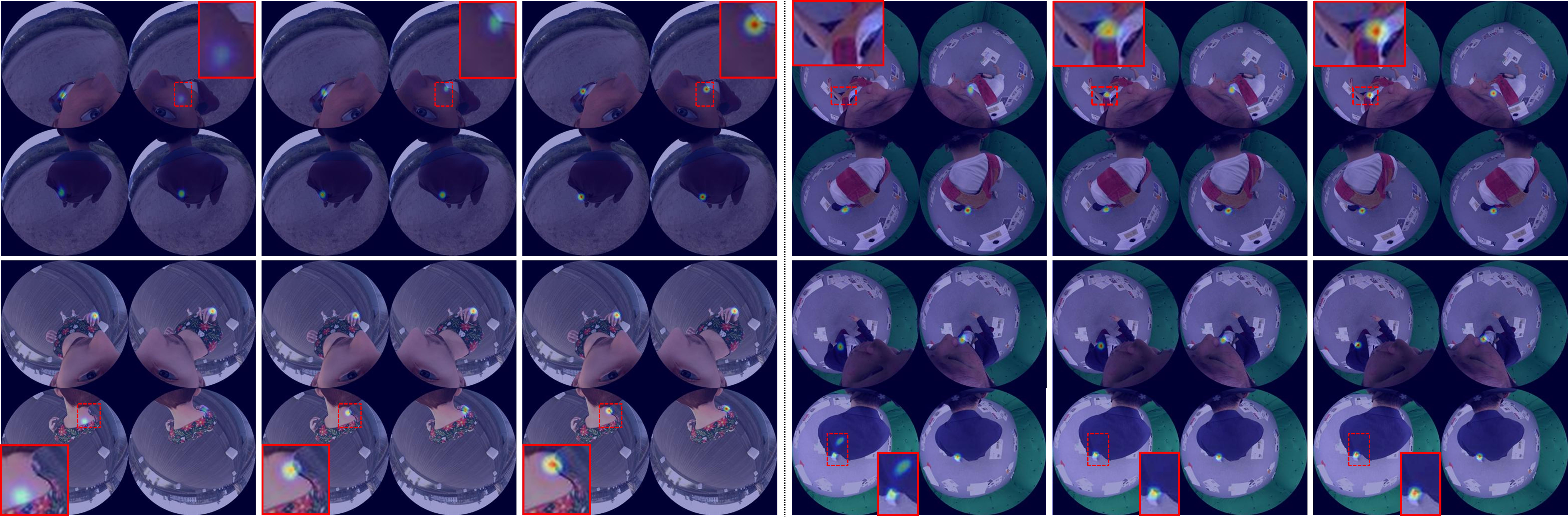}
    \vspace{-7mm}
    \caption*{\footnotesize{
        Initial heatmaps
        \,\,\,\,\,\,\,\,\,\,\,\,\,\,\,\,\,\,
        Refined heatmaps
        \,\,\,\,\,\,\,\,\,\,\,\,\,\,\,\,\,\,\,\,\,
        Ground truths
        \,\,\,\,\,\,\,\,\,\,\,\,\,\,\,\,\,\,\,\,\,\,\,\,\,\,
        Initial heatmaps
        \,\,\,\,\,\,\,\,\,\,\,\,\,\,\,\,\,\,
        Refined heatmaps
        \,\,\,\,\,\,\,\,\,\,\,\,\,\,\,\,\,\,\,\,
        Ground truths
        \,\,
    }}
    \vspace{-3mm}
 \caption{\textbf{Visualizations of 2D heatmap refinement with our method.} \textbf{Left}: Ego4View-Syn. \textbf{Right}: Ego4View-RW. We highlight notably refined regions with red bounding boxes and provide zoomed-in views of these areas for closer examination.
 }
 \label{fig:qualitaitve_result_hm2d}
\vspace{-4mm}
\end{figure*}

\begin{table}[t]
\centering
\scalebox{0.7}{
\begin{tabular}{lccccc}
\hline
\noalign{\smallskip}
\multirow{2}{*}{Method (4 views)} & \multicolumn{2}{c}{2D Heatmap} & \multicolumn{2}{c}{3D Pose}   \\ 
                        & Front MSE & Back MSE           & MPJPE & PA-MPJPE \\ 

\noalign{\smallskip}
\hline
\noalign{\smallskip}
Baseline~\cite{yang2024egoposeformer} & 2.30 & 3.36 & 20.20 & 17.63 \\

\noalign{\smallskip}
\cdashline{1-5}
\noalign{\smallskip}

(a) no 2D anchor & 2.10 & 2.76 & 20.05 & 17.71 \\
(b) no HM uncertainty & \bf 1.67 & 1.93 & 19.67 & 17.26 \\
(c) no RGB \& HM embed & 1.78 & \underline{1.84} & 19.47 & \underline{17.17} \\
(d) no RGB embed & 1.69 & 1.87 & \underline{19.45}  & 17.22 \\

\noalign{\smallskip}
\cdashline{1-5}
\noalign{\smallskip}

Ours & \underline{1.68} & \bf 1.83 & \bf 19.25 & \bf 16.95 \\

\noalign{\smallskip}
\hline
\end{tabular}
}
\vspace{-3mm}
\caption{\textbf{Ablation study of our method on Ego4View-Syn} with 2 front and 2 rear views. \textbf{MSE}: mean squared error ($\times 10^{-4} / \text{pixel}$).}
\label{table:ablation_method}
\vspace{-5mm}
\end{table}

\noindent\textbf{Ablation study.} 
In Table~\ref{table:ablation_method}, we first ablate our refinement module (a) with standard cross attention (no 2D anchor) instead of deformable attention. 
This variant struggles to effectively refine 2D heatmaps, leading to less accurate 3D pose tracking than our complete model. 
Next, we omit the uncertainty-aware masking from our method (b), which results in diminished 2D heatmap accuracy in rear views and a noticeable decline in 3D pose estimation performance. 
We hypothesize that this decline stems from substantial self-occlusions in rear views that the model cannot adequately address without uncertainty-aware masking. 
We further ablate our method by excluding either the initial heatmap embeddings or the heatmap and RGB image embeddings jointly. 
In both cases, the estimation accuracy decreases. 

Fig.~\ref{fig:qualitaitve_result_hm2d} visualizes the results of our refinement process with significantly improved regions marked by red boxes. 
The left parts of Fig.~\ref{fig:qualitaitve_result_hm2d} show heatmap values for the hands on Ego4View-Syn. 
While the initial heatmaps highlight incorrect regions, our refinement module improves the localization accuracy substantially. 
Notably, although the right hand is visible in the rear views in Fig.~\ref{fig:qualitaitve_result_hm2d}-(bottom left), the initial heatmaps exhibit erroneous regions with high confidence. 
Our method, however, accurately offsets these predictions. 
Similarly, the right parts of Fig.~\ref{fig:qualitaitve_result_hm2d} display heatmap regions for the left foot on Ego4View-RW. 
The initial estimates are either missing or incorrectly detect multiple locations. 
We attribute the latter case to a potential network bias towards rear views, where lower body parts often appear beneath the upper body (\eg due to the forward positioning of the human head relative to the skeleton center). 

Despite these challenges, our refinement approach yields reliable results.
The improved 2D joint estimation holds significant value in advancing downstream applications.
As demonstrated in prior works, refined 2D joint locations could enhance the performance of foundation models~\cite{kirillov2023segment} for generating accurate body masks~\cite{hakada2024unrealego2} and serve as critical inputs for motion priors, facilitating improved motion generation~\cite{ wang2024egocentric} and optimization processes~\cite{wang2021estimating}.

%% file: sec/6_conclusion.tex
\noindent\textbf{Limitations.} 
While this paper highlights multiple advantages of an HMD with rear cameras, the setup can be less effective for users with frizzy or voluminous hair that can obstruct the field of view of the cameras. 
Also---although our approach significantly improves the 3D accuracy on average---it still struggles to estimate physically plausible 3D poses under conditions of extreme self-occlusion (\eg at 6:29 in our supplementary video); incorporation of physics-based motion priors could be investigated in future.

\section{Conclusion} 
We studied for the first time the potential of rear cameras for egocentric 3D full-body tracking and proposed a new transformer-based method to address the challenges associated with the integration of rear views (\ie unreliable 2D joint detection).
Our experiments suggest that although the inclusion of rear cameras enhances average pose tracking accuracy, naively incorporating rear views alongside frontal inputs does not always achieve optimal estimation for existing methods and occasionally degrades their accuracy. 
In contrast, the proposed method addresses this challenge with multi-view context, achieving state-of-the-art results, \eg ${>}10\%$ gains on Ego4View-RW. 
Also, our ablation shows that our deformable attention mechanism, guided by estimated 2D joints, substantially facilitates the extraction of multi-view heatmap features.
Furthermore, our novel large-scale datasets provide valuable annotations for egocentric 3D human pose estimation with rear views.
Our work will encourage the community to rethink innovative HMD designs tailored for 3D full-body tracking.
The HMD designs with rear cameras will open up possibilities for many future works, \eg~not only pose tracking but also body shape and avatar reconstruction as well as human-object interaction.

\noindent\textbf{Acknowledgment.} 
Hiroyasu Akada was supported by the Nakajima Foundation scholarship.

%% file: sec/X_suppl.tex
This supplementary material provides more details about our work. 
\textbf{Please also watch our supplementary video for dynamic visualizations, including the proposed datasets, qualitative results, and character animations as a future application of our method}.

\section{Per-Action Evaluation}
Table~\ref{table:quantitative_result_rw_action} presents the 3D error evaluation for each action category on Ego4View-RW.
Our refinement method consistently brings substantial improvement across all action types.
Notably, our proposed module demonstrates superior effectiveness, particularly for motions involving lower-body movements, achieving a 15.5\% improvement in ``stretching legs''.
This enhancement is likely attributed to the frequent self-occlusion of the legs in either front or rear views, which often leads to misrepresentation in joint detection by the current state-of-the-art method~\cite{yang2024egoposeformer} even with four-view inputs.
In contrast, our module effectively mitigates these challenges, resulting in significant advancements in egocentric 3D human pose estimation with rear-view integration.

\section{Joint Visibility Calculation}
As described in Sec.~3, we obtain the visibility of end-effector joints (hands and feet) in our synthetic setup.
We first generate 2D egocentric fisheye views using SMPL models with the predefined body part segmentation mesh~\cite{SMPL:2015}.
Next, we project ground-truth 3D joints onto these images to obtain reference points, querying the nearest 2D points within a $10\times10$ pixel region around each reference.
We classify a 3D joint as visible if any queried 2D point corresponds to its respective body part; otherwise, it is considered occluded.

\section{Additional Details of Network Architecture}
As mentioned in Sec.~4, we use several shallow networks, \ie $\mathcal{F}_{\text{O}}$, $\mathcal{F}_{\text{R}}$, $\mathcal{F}_{\text{HM}}$, $\mathcal{P}_{\text{HM}}$, $\mathcal{P}_{\text{RGB}}$, and $\mathcal{P}_{\text{Q}}$.
$\mathcal{F}_{\text{O}}$ and $\mathcal{F}_{\text{R}}$ consist of two linear layers with a bilinear up-sample operation as well as with an intermediate dimension size of 64 and 128, respectively.
$\mathcal{F}_{\text{HM}}$ uses one convolutional layer with a kernel size of 3, a stride of 2, and a padding size of 1, followed by two linear layers with an intermediate dimension size of 256, a bilinear up-sample operation, and a linear layer to generate heatmaps.
$\mathcal{P}_{\text{HM}}$ is composed of two linear layers with intermediate and output dimension sizes of 256 whereas $\mathcal{P}_{\text{RGB}}$ and $\mathcal{P}_{\text{Q}}$ are a single liner layer with output dimension size of 256.

\begin{table}[t]
\centering
\scalebox{0.81}{
\begin{tabular}{lccccc}
\noalign{\smallskip}
\hline
\noalign{\smallskip}
\multicolumn{1}{l}{Method} & walking &  kicking & boxing & crouching   \\ 
\noalign{\smallskip}
\hline
\noalign{\smallskip}

EPF~\cite{yang2024egoposeformer}  

& 45.16
& 74.50
& 71.22
& 50.89 \\

\,\, + Ours

& \bf 42.10
& \bf 68.85
& \bf 60.81
& \bf 46.20 \\

\noalign{\smallskip}
\hline
\hline
\noalign{\smallskip}
\multicolumn{1}{l}{Method} & kneeing & crawling & dancing &  \begin{tabular}[c]{@{}c@{}} twisting \\ body \end{tabular} \\ 
\noalign{\smallskip}
\hline
\noalign{\smallskip}

EPF~\cite{yang2024egoposeformer}  

& 87.14
& 82.35
& 52.99
& 61.64 \\

\,\, + Ours  

& \bf 73.69
& \bf 76.92
& \bf 48.75
& \bf 56.03 \\

\noalign{\smallskip}
\hline
\hline
\noalign{\smallskip}
\multicolumn{1}{l}{Method} & \begin{tabular}[c]{@{}c@{}} stretching \\ arms \end{tabular} & \begin{tabular}[c]{@{}c@{}} stretching \\ legs  \end{tabular}  & \begin{tabular}[c]{@{}c@{}} rotating \\ shoulders \end{tabular}  & \begin{tabular}[c]{@{}c@{}} raising \\ legs \end{tabular}  \\ 
\noalign{\smallskip}
\hline
\noalign{\smallskip}

EPF~\cite{yang2024egoposeformer}  

& 50.03
& 58.14
& 53.56
& 75.62 \\

\,\, + Ours  

& \bf 46.88
& \bf 49.09
& \bf 51.94
& \bf 68.91 \\

\noalign{\smallskip}
\hline
\hline
\noalign{\smallskip}
\multicolumn{1}{l}{Method} & \begin{tabular}[c]{@{}c@{}} balancing \\ legs up behind  \end{tabular} &  \begin{tabular}[c]{@{}c@{}} sitting \\ on the ground  \end{tabular} &  & all \\ 
\noalign{\smallskip}
\hline
\noalign{\smallskip}

EPF~\cite{yang2024egoposeformer}  

& 82.82
& 77.19
& 
& 63.38 \\

\,\, + Ours  

& \bf 74.24
& \bf 66.79
& \bf 
& \bf 56.94 \\

\noalign{\smallskip}
\hline
\noalign{\smallskip}
\end{tabular}
}
\vspace{-3mm}
\caption{\textbf{Per-action evaluation on Ego4View-RW} (MPJPE) with 2 front and 2 rear views. EPF represents EgoPoseFormer~\cite{yang2024egoposeformer}.
}
\label{table:quantitative_result_rw_action}
\vspace{-4mm}
\end{table}

\section{Rear-View Integration for Existing Method}
As mentioned in Secs.~4 and 5, the refined heatmap features  $\mathbf{R}_{k}$ and heatmaps $\widetilde{\mathbf{H}}_{k}$ can be utilised with existing 2D-to-3D lifting modules to estimate 3D poses.
In this work, we integrate our module into the current state-of-the-art methods, EgoPoseFormer~\cite{yang2024egoposeformer} and EgoTAP~\cite{kang2024egotap}.
However, unlike the 3D module of EgoTAP~\cite{kang2024egotap} that directly uses heatmaps as inputs, EgoPoseFormer~\cite{yang2024egoposeformer} uses the heatmap features (before the final heatmap output layer) instead of the heatmaps in their 2D-to-3D lifting module.
To account for their methodology, we first input the refined features $\mathbf{R}_{k}$ from all views into a simple network consisting of four convolutional layers followed by a linear layer, yielding an initial 3D pose $\mathbf{P} \in \mathbb{R}^{16 \times 3}$, which represents 16 joints including the head.
This initial 3D pose is subsequently fed into the 3D updating module of EgoPoseFormer~\cite{yang2024egoposeformer} to produce an updated 3D pose $\mathbf{P}_{\text{final}}$ as the final output.

\section{Model Size}
Here, we provide the details of our model size and inference speed with 2 front and 2 rear views.
As mentioned in Sec. 4 and 5, we adopt the current state-of-the-art methods as a baseline, \ie EgoPoseFormer~\cite{yang2024egoposeformer} (27M parameters) and EgoTAP~\cite{kang2024egotap} (242M parameters), with our refinement module (25M $\times$ the number of views).
Note that EgoPoseFormer~\cite{yang2024egoposeformer} consists of a simple UNet-based architecture and three transformer-based layers; therefore, they tend to have fewer trainable parameters than EgoTAP~\cite{kang2024egotap}.

Regarding the inference speed on our setup with a single NVIDIA 
Quadro RTX 8000 and PyTorch, while the original EgoPoseFormer~\cite{yang2024egoposeformer} runs at 67 fps, EgoPoseFormer with our refinement module can run at 30 fps with 2 front and 2 rear views. Therefore, our method can be utilized with real-time applications. 
Note that our focus lies in making the best use of rear views and prioritizes tracking accuracy; future work will improve inference speed with our proposed setup and novel large-scale datasets, \ie Ego4View-Syn and Ego4View-RW. 